\documentclass[letterpaper, 10 pt, conference]{ieeeconf}
\IEEEoverridecommandlockouts 
\makeatletter
\let\NAT@parse\undefined
\makeatother
\usepackage[numbers]{natbib}

\usepackage{booktabs}
\usepackage{multirow}
\usepackage{amssymb}
\usepackage{amsmath}

\usepackage{hyperref}

\usepackage{comment}

\usepackage{graphicx}
\graphicspath{ {./images/} }

\usepackage{caption}
\usepackage{subcaption}

\title{\LARGE \bf
Learning to Detect Multi-Modal Grasps \\ for Dexterous Grasping in Dense Clutter
}

\author{Matt Corsaro$^{1}$, Stefanie Tellex$^{1}$, George Konidaris$^{1}$%
\thanks{$^{1}$Department of Computer Science, Brown University
        {\tt\small \{mcorsaro, stefie10, gdk\}@cs.brown.edu}}%
}

\begin{document}

\maketitle
\thispagestyle{empty}
\pagestyle{empty}

\begin{abstract}
We propose an approach to multi-modal grasp detection that jointly predicts the probabilities that several types of grasps succeed at a given grasp pose.
Given a partial point cloud of a scene, the algorithm proposes a set of feasible grasp candidates, then estimates the probabilities that a grasp of each type would succeed at each candidate pose.
Predicting grasp success probabilities directly from point clouds makes our approach agnostic to the number and placement of depth sensors at execution time.
We evaluate our system both in simulation and on a real robot with a Robotiq 3-Finger Adaptive Gripper and compare our network against several baselines that perform fewer types of grasps.
Our experiments show that a system that explicitly models grasp type achieves an object retrieval rate 8.5\% higher in a complex cluttered environment than our highest-performing baseline.
\end{abstract}

\section{Introduction}
\label{sec:intro}
Grasping is one of the most important open problems in robotics---it is the most essential skill for pick-and-place tasks and a prerequisite for many other manipulation tasks, such as tool use \citep{george_survey}.
If robots are to one day perform the complex manipulation tasks humans are capable of in varying home and workplace environments, they must first master grasping.

\begin{figure*}
    \vspace{7pt}
    \centering
    \begin{subfigure}[b]{0.18\textwidth}
        \centering
        \includegraphics[width=\textwidth]{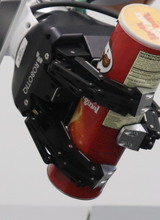}
        \caption{Wide Power}
        \label{fig:wide_pow}
    \end{subfigure}
    \hfill
    \begin{subfigure}[b]{0.18\textwidth}
        \centering
        \includegraphics[width=\textwidth]{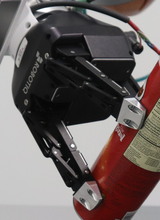}
        \caption{Wide Precision}
        \label{fig:wide_prec}
    \end{subfigure}
    \hfill
    \begin{subfigure}[b]{0.18\textwidth}
        \centering
        \includegraphics[width=\textwidth]{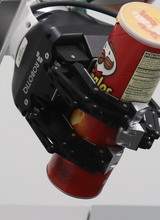}
        \caption{Basic Power}
        \label{fig:basic_pow}
    \end{subfigure}
    \hfill
    \begin{subfigure}[b]{0.18\textwidth}
        \centering
        \includegraphics[width=\textwidth]{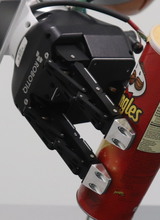}
        \caption{Basic Precision}
        \label{fig:basic_prec}
    \end{subfigure}
    \hfill
    \begin{subfigure}[b]{0.18\textwidth}
        \centering
        \includegraphics[width=\textwidth]{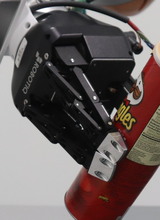}
        \caption{Pincher}
        \label{fig:pincher}
    \end{subfigure}
    \hfill
    \caption{The five main grasp types achievable using the Robotiq 3-Finger Adaptive Gripper. Our system detects a grasp type and pose that would lead to a successful grasp, enabling it to more effectively clear clutter.}
    \label{fig:grasp_types_pringles}
\end{figure*}

Humans use multiple types of grasps, depending on the object, the task, and the scene \citep{dollarkragic}.
A human may perform a large-diameter power grasp to stably grasp the handle of a heavy jug, but a precision sphere grasp to lift a golf ball off the ground.
If the clutter around an object precludes one particular grasp type, humans simply switch to another.
It is therefore natural that the ability to use multiple grasp modalities would substantially improve robots' ability to grasp a wide range of objects, especially in dense clutter.
However, state-of-the-art grasp detection systems typically detect pincher grasps exclusively, and are evaluated using small objects and two-finger parallel-jaw grippers \cite{rob-journal, dexnet2, dieter-collision, qtopt}.
Existing grippers are capable of executing multi-finger dexterous grasps better suited  to stably grasping both small and larger, heavier objects; the grasps a Robotiq 3-Finger Adaptive Gripper is capable of executing are demonstrated in Figure \ref{fig:grasp_types_pringles}.
Several grasp detection approaches are applicable to multi-finger grippers, but are only capable of performing one type of grasp \citep{rss2020}, or do not explicitly model grasp type~\citep{kappler}.
Some have taken grasp type into consideration, but are evaluated on singulated objects \cite{tucker-grasp-types}, rely on human-labeled data \cite{human-label-grasp-types}, or return fingertip placement for fully actuated fingers \cite{varley}.
Furthermore, these systems are not evaluated in dense clutter.

We propose a data-driven grasp detection framework that, given partial depth data and a grasp pose, jointly predicts the grasp success probabilities of several types of grasps.
We train a deep neural network to perform this joint classification using a dataset containing grasp candidates generated from real point clouds and grasp labels generated in simulation.
Given a point cloud---captured from an arbitrary number of depth sensors in arbitrary poses---along with a grasp pose, our network outputs a probability for each available grasp modality.
These values reflect the probability that the corresponding type of grasp would succeed at the given  pose.

We evaluate our system both in simulation and experimentally on a Robotiq 3-Finger Adaptive Gripper.
We first evaluate our system on a held-out test set from simulated data to show that our network efficiently learns to jointly predict grasp type when compared to a larger ensemble of networks.
On a real robot, our system clears objects from cluttered tabletop piles containing objects of varying sizes.
To show the usefulness of multiple grasp modalities in dense clutter, we compare against several ablations of our network capable of performing fewer grasp types, and find that a system capable of multiple grasp types clears more objects than baselines that use fewer.

\section{Background}

With recent advances in deep learning, data-driven grasp detectors have proven effective for generating parallel-jaw grasps for two-finger grippers.
Given visual information, these systems return an end effector pose at which an executed grasp would likely be successful.
Most state-of-the-art parallel-jaw grasp detectors, such as \citet{rob-journal} and \citet{dexnet2}, follow a two-step proposal-evaluation model.
A proposal function $\mathbf{PROP}: P \rightarrow G$, implemented as a heuristic \citep{rob-journal} or a generative network \citep{nvidia-vae}, first generates a large set of 6-DoF end effector poses $G \subseteq \mathbb{SE}(3)$ from a point cloud $P \subseteq \mathbb{R}^3$.
A grasp evaluation neural network $\mathbf{EVAL}: g \in G \rightarrow [0, 1]$ then maps each $g$ to a probability.

Another common approach is to train a neural network to predict optimal actions using a reinforcement-learning framework.
Works such as \citet{qtopt} and \citet{handeye} train their systems using real robot data, which is time-consuming to produce.
Though such systems can achieve state-of-the-art grasp success rates, their reliance on reinforcement learning makes them brittle; the same camera configuration used while training is required at test time.
Furthermore, modifying the system to grasp a specified object is not straightforward as it is in proposal-evaluation systems, where the proposal step can be easily modified without adjusting a reward function or retraining.
Though both of these types of systems enable two-finger parallel-jaw grippers to grasp some objects, these grippers are capable of executing only simple pincher grasps.

Data-driven grasp detection frameworks have also been applied to perform multi-finger dexterous grasping.
However, these systems are either capable of performing only fingertip or precision grasps \cite{unigrasp, rss2020}, or use supervised \citep{kappler, tucker, schmidt, song, liu2019iros} or reinforcement learning \citep{bohan} to evaluate or predict wrist poses and finger pre-grasp parameters, but do not explicitly model grasp type.

A few recent works predict grasp stability for multiple grasp types.
\citet{tucker-grasp-types} train two classifiers to predict power and precision success probabilities from a shared embedding for a 4-finger Allegro Hand.
Each classifier is evaluated separately on singulated objects on which power grasps are always preferred.
As our system jointly classifies candidates of each grasp type at a given position, it returns both a predicted optimal grasp pose and type.
We evaluate our system in a cluttered real-world scenario where multiple types of grasps can be necessary to clear the scene.
\citet{human-label-grasp-types} and \citet{icra_review_suggestion} use human-labeled data to train a neural network to predict grasp type, while our system learns to use grasp types from simulated grasping data, avoiding human bias or error.
Both systems are not evaluated in dense clutter.
\citet{varley} employ a hybrid approach, using a deep neural network to guide fingertip placement and a grasp planning simulator to localize gripper placement.
They define a set of canonical grasp types based on the most common finger pre-poses in their simulated training set.
Planning fingertip placement can be difficult for underactuated grippers; to perform a power grasp with an underactuated gripper, each finger's more proximal links would make contact with the object first, making the final distal link placement less relevant.
Their system is also not evaluated in dense clutter.
\citet{hierarchical-type} use hierarchical reinforcement learning to select a grasp type and grasp location.
They maintain a dataset of successful grasps for each grasp type and match new point clouds to this dataset using ICP.
Their system is not evaluated in clutter.
As they employ a reinforcement-learning framework, a higher-level controller could not request a grasp type or target object as it could with ours.
Our approach is capable of executing multiple grasp types, allowing it to successfully grasp objects in a variety of real-world, cluttered experimental scenarios.

\section{Learning to Detect Multi-Modal Grasps}
\label{sec:approach}
Though parallel-jaw grasp detectors have proven successful, two-finger grasps can be insufficient when a robot deals with large, heavy objects.
Grasp detectors designed for multi-finger grippers detect grasps of a single type, and those that can explicitly utilize multiple grasp types have not been proven to enable a robot to clear a pile of dense clutter.
Our system demonstrates the usefulness of multiple grasp types when picking objects of varying sizes from piles of dense clutter using the proposal-evaluation paradigm commonly used in grasp detection systems \cite{rob-journal}.
The proposal function $\mathbf{PROP}: P \rightarrow G$ generates a set of 6-DoF end effector poses $G \subseteq \mathbb{SE}(3)$ from a partial point cloud $P \subseteq \mathbb{R}^3$.
Unlike the grasp evaluators $\mathbf{EVAL}: g \in G \rightarrow [0, 1]$ used in related works that map a grasp pose to a single probability \cite{rob-journal, dexnet2}, our grasp evaluation neural network $\mathbf{EVAL}: g \in G \rightarrow [0, 1]^n$ maps each $g$ to a vector of $n$ success probabilities, each corresponding to a different grasp type.
This architecture enables us to jointly predict the probabilities of success for multiple grasp types at a given $g$.

We generate grasp pose candidates $G \subseteq \mathbb{SE}(3)$ using the 6-DoF candidate generation algorithm $ \mathbf{GEN}: P \rightarrow G $ proposed by \citet{rob-candidate}: given a point cloud of an object or cluttered pile of objects represented as a set of 3D points $P \subseteq \mathbb{R}^3$, sample a subset $C \subseteq P$ of $k$ grasp candidate centroid positions.
Each $c_s \in C$ is assigned a single orientation $o_s \in \mathbb{SO}(3)$ based on the normals and curvature estimated at $c_s$; the gripper approach direction is anti-parallel to the estimated normal, and the gripper closes along the curvature.
Similar candidates can be sampled by rotating the sampled orientation about the approach direction; we rotate by $90^{\circ}$ to generate one additional pose.
Finally, candidates causing the gripper to collide with $P$ are pruned.
The candidate generation algorithm returns a set of $k$ proposed candidates $\mathbf{GEN}(P) = G$ where $g_s \in G$ and $g_s = \{c_s, o_s\}$.

\begin{figure*}
\vspace{7pt}
\centerline{\includegraphics[scale=0.5]{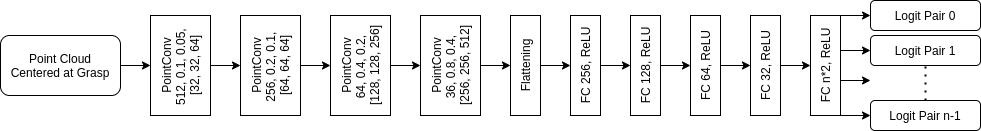}}
\caption{Network diagram of $\mathbf{DNN}$. PointConv layer parameters listed are number of points, radius, sigma, and MLP sizes. Fully connected layer parameters are output size.}
\label{fig:net_diag}
\end{figure*}

The second phase of our system evaluates each of the $k$ proposed candidates $g_s \in G$.
A deep neural network estimates success probabilities for each grasp type at each $g_s$, taking $P$ and an encoding of $g_s$ as input.
As several recent papers have shown \cite{tucker-completion, pointnetgpd}, grasp candidates can be efficiently encoded directly from point clouds recorded from arbitrary viewpoints using a PointNet-inspired architecture \cite{pointnet}.
The encoding layers used in our network are based on the PointConv architecture \citep{pointconv}.
We encode a candidate grasp pose by centering $P$ at $c_s$ and aligning $P$'s orientation with $o_s$.
We then crop all points outside the approximate grasping region, represented as a box around the fingers and the area they sweep through.
This transformation $\mathbf{TF}: P, g_s \rightarrow P_s$ produces $P_s$, an encoding of a grasp pose and the object geometry local to it.
This transformed, cropped cloud $P_s$ representing a single $g_s$ is then fed to the network $\mathbf{DNN}$, which is illustrated in Figure \ref{fig:net_diag}.

The encoding layers in our network consist of four PointConv feature encoding layers.
Following \citet{tucker-completion}, we reduce the first layer's number of points from 1024 to 512 and the third layer's final multi-layer perception from 128 to 64 units.
The output from the fourth encoding layer is then fed through a series of five fully connected layers with ReLU activations.
The final fully connected layer outputs a logit pair for each of the $n$ grasp types the gripper is capable of:
$\mathbf{DNN}: P_s \rightarrow X_s \in \mathbb{R}^{n \times 2}$.
We output two logits per grasp type in order to train the network to perform binary classification on each grasp type and predict whether a grasp of each type would succeed or fail.
These logit pairs are passed through $n$ independent softmax functions.
The $n$ resulting probabilities corresponding to positive labels, $\sigma(X_s)_{*,1} = s_s \in [0,1]^n$, can be interpreted as the probabilities that a grasp at $g_s$ of the corresponding grasp type would succeed.

We train $\mathbf{DNN}$ to jointly perform $n$ binary classifications using a summed cross entropy loss function.
Joint binary classification is useful in cases where multiple entangled predictions are made from a single input source.
By training a single network to perform joint binary classification, our system learns an embedding that efficiently encodes the information required to determine whether each grasp type succeeds given a cloud and grasp pose.
Though joint binary classification has been proposed to solve problems such as emotion detection \citep{emotion}, ours is the first robotics application we are aware of that uses it.
We define this summed cross entropy loss function (equation \ref{eq:loss}) as a modified form of the standard cross-entropy loss function for $m$-class classification,
\begin{equation}
-\sum\limits_{c=0}^{m-1}y_c~log(b_c),
\label{eq:loss_vanilla}
\end{equation}
where $y_c$ is 1 if $c$ is the correct label for a given exemplar and 0 otherwise and $b_c$ is the estimated probability that the exemplar is of class $c$.
Given a labeled grasp exemplar $e = \{g, P, l\}$ where $l \in \{0,1\}^n$ and $P_g = \mathbf{TF}(P, g)$, we compute the summed cross entropy between $l$ and $\sigma(\mathbf{DNN}(P_g)) = Z \in [0,1]^{n \times 2}$.
The summed cross entropy loss for our joint binary classification problem is:

\begin{equation}
-\sum\limits_{i=0}^{n-1}\sum\limits_{c=0}^{1}y_{i,c}~log(Z_{i,c}),
\label{eq:loss}
\end{equation}
where $y_{i,c}$ is 1 if $c=l_i$ and 0 otherwise.
Training details are found in Section \ref{sec:training}.

In our experiments, given some $P$ of an object or a set of objects, we generate a set of grasp candidates $G = \mathbf{GEN}(P, k)$ where $|G|=k$.
With $\mathbf{DNN}$ trained to evaluate grasps for a specific gripper, we predict each candidate success probability vector $s_s = \sigma(\mathbf{DNN}(\mathbf{TF}(P, g_s)))_{*,1}~\forall g_s \in G$ to get $S \in [0,1]^{k \times n}$ where $s_s \in S$.
Finally, we select the grasp pose $g_m = \{c_m, o_m\}$ and grasp type $i_m$ corresponding to the maximum entry in $S$ that is collision free.
When executing this grasp, the gripper is first moved to a pre-grasp pose some distance $d$ away from $\{p_m, o_m\}$ along the negative approach direction.
Finally, the gripper is moved to $\{p_m, o_m\}$ and the fingers are closed to complete the grasp.

\subsection{Dataset Generation \& Network Training}
\label{sec:training}
In order to train $\mathbf{DNN}$, a dataset of grasp exemplars $E$ where $e = \{g, P, l\} \in E$ and $l \in \{0,1\}^n$ is required.
The BigBIRD dataset \citep{bigbird} contains a set of real partial point clouds captured from 600 viewpoints on a set of common household products and a complete mesh for each object.
BigBIRD is a popular dataset for training grasp detection systems since no simulation-to-real transfer is required with real point clouds.
Our grasp dataset is generated from 20 BigBIRD objects and 12,000 point clouds.
We generate a set of grasp candidates $G$ from these clouds using $GEN$.

Each candidate is then assigned a label $l \in \{0,1\}^n$.
A label is generated by attempting each grasp type $i$ at $g$ in simulation and recording whether or not the grasp succeeds.
We perform grasps in the Drake simulation environment \citep{drake} to accurately simulate the Robotiq 3-Finger Adaptive Gripper and its grasp types.
The simulated graspable object models are based on the complete BigBIRD object models.
Each object model is placed on a plane, the gripper model is placed at the grasp pose specified by the sampled candidate $g$ and grasp type $i$.
After the gripper executes a grasp on the object and the plane is removed from the scene, $l_i$ is set to 1 if the object remains between the gripper's fingers or 0 if it falls out of the grasp.
Candidate grasps at which no grasp type cause a collision between the gripper and object or table before execution are kept in the dataset, and those that cause collisions are removed.
The resulting dataset $E$ contains over 36,000 grasp candidates, each assigned labels $l$ generated by simulating over 180,000 grasp attempts using $n=5$ grasp types.
This dataset is not perfectly balanced between positive and negative labels; the percentages of positive labels for each grasp type are 79\%, 43\%, 81\%, 73\%, and 58\%.
Though our labels are generated on singulated objects, \citet{rob-journal} showed that systems trained with grasps on singular objects generalize well to real-world clutter.

The network architecture described in Section \ref{sec:approach} is then trained to perform joint binary classification using the summed cross entropy loss function defined in equation \ref{eq:loss}.
This network is implemented in TensorFlow and trained on a single GeForce GTX 1080 Ti.
We use a learning rate of $10^{-5}$, a batch size of 16, and the Adam optimizer to train the network.
Like \citet{rob-journal}, the cloud input to our network is comprised of two point clouds; one cloud is the BigBIRD cloud used to generate a candidate, and the second is the cloud captured from the same viewing angle $54^{\circ}$ away from the first.
As the BigBIRD dataset is generated by capturing point clouds of objects from five fixed viewing angles and rotating the object in $3^{\circ}$ increments on a turntable, these secondary clouds are available in the BigBIRD dataset.
Since our network takes transformed and cropped point clouds $P_s$ as input and classifies 6-DoF grasp candidates, the number of point clouds captured and their viewing angles are arbitrary.
We choose to capture two point clouds at training and test time to sufficiently cover the workspace when generating grasp representations $P_s$, but the configurations need not be the same.

\subsection{Grasp Types}
\label{sec:grasp_types}
A grasp taxonomy for any multi-finger robotic gripper could be derived to determine the number of grasp types $n$ that it is capable of, much like the one \citet{dollarkragic} present to categorize 33 grasp types humans are capable of.
Though our framework is compatible with any robot gripper and its $n$, we train and evaluate our system with the Robotiq 3-Finger Adaptive Gripper, a 4-DoF, 11-jointed underactuated gripper.
Unlike the fully articulated grippers used in related work \citep{tucker-grasp-types}, the Robotiq 3-Finger Adaptive Gripper is designed specifically to perform different types of grasps \cite{robotiq}.
Each 3-jointed finger is controlled by one motor, and an additional actuator adjusts the orientation of the non-thumb fingers.
To avoid self-collision, the gripper's controller allows for three discrete operating modes.
The non-thumb fingers are parallel in \textbf{basic} mode, spread apart in \textbf{wide} mode, and brought together in \textbf{pincher} mode.
The gripper is capable of performing two types of grips: a \textbf{fingertip} or precision grip occurs when the distal links grip an object, while an \textbf{encompassing} or power grip occurs when the proximal links first contact an object, causing the fingers to wrap around it.
The grip executed depends on the distance from the gripper's palm to the target object.
In basic and wide mode, both encompassing and fingertip grips are possible, while in pincher mode, fingertip grips emulate parallel-jaw grippers.
These operating modes and grip types enable the gripper to execute five types of grasps, illustrated in Figure \ref{fig:grasp_types_pringles}: \textbf{basic power} and \textbf{wide power} are useful for firmly grasping large objects, \textbf{basic precision} and \textbf{wide precision} can grasp objects from a surface, and a precision \textbf{pincher} that emulates a parallel-jaw gripper is useful for precise tabletop grasps on small objects.

The mechanical design of the Robotiq 3-Finger Adaptive Gripper enables it to execute multiple types of grasps with a single simple control policy without the need for tactile sensors or joint encoders.
It enables us to parameterize grasp type over the operating mode and distance from the object to the gripper.
To execute a grasp of a given type $i_m$ at candidate $g_m = {c_m, o_m}$, we assign the gripper a pose $\{p_m, o_m\}$.
We define $p_m$ to be the position of the point at the center of the gripper's palm.
The orientation of this pose is the same as the candidate's pose.
As $c_m$ is a point sampled from the cloud $P$, assigning $p_m=c_m$ would result in a collision.
$p_m$ is instead set some distance $d$ away from $c_m$ along the negative approach direction depending on the grip type required to execute a grasp of type $i_m$: $p_m = c_m - o_m d$.
To execute an encompassing grasp, $p_m$ should be close to $c_m$ so the fingers' proximal links, first make contact with the object and wrap around it.
$p_m$ should be sufficiently far from $c_m$ during a fingertip grasp so the fingers' distal links would likely make contact with the object.
We therefore set $d_e=1.9$ cm to perform an encompassing grip when executing a basic or wide power grasp, and set $d_f=8.22$ cm to perform a fingertip grip when executing a basic precision, wide precision, or pincher grasp.

\section{Measuring Network Generalization}
\label{sec:sim_exp}
We first evaluate our system by testing its performance on several held-out datasets.
In the first scenario, the test set is comprised of 15\% of the grasp candidates in our dataset, selected at random, while the remaining 85\% are used to train the network.
The second, more difficult scenario tests how well the system generalizes to unseen objects.
Here, the test set contains all grasp candidates from 15\% of the objects in the dataset, while the grasps on the remaining objects are used to train the network.
In each scenario, we compare our architecture that outputs $n=5$ predictions per pose from a shared embedding (COMBINED) to a similar architecture that uses an ensemble of $n$ individual deep networks to predict grasp success for each grasp type (SEPARATE).
These individual networks are a naive approach that uses $n$ times as many parameters as the combined network, but provide an upper bound to compare our system against.
Table \ref{tab:simacc} shows the test-set classification accuracies of our system and the baseline trained and tested using the two dataset divisions.
These results are each averaged over three random seeds used to divide the dataset.

\begin{table*}[ht]
\centering
\begin{tabular}{ccccccccccc}
\\
\toprule
Split & Arch. & \# Params & Avg Acc & Avg Prec & Avg F1 & T1 Acc & T2 Acc & T3 Acc & T4 Acc & T5 Acc \\
\midrule
& COMBINED & {10.4M} & 0.867 & {0.879} & 0.897 & 0.913 & 0.802 & {0.908} & 0.824 & {0.886} \\ 
\multirow{2}{*}[+10pt]{Rand.} & SEPARATE & 51.9M & {0.871} & 0.878 & {0.9} & {0.922} & {0.804} & {0.908} & {0.833} & {0.886} \\
\midrule
& COMBINED & {10.4M} & 0.829 & 0.869 & 0.881 & 0.841 & 0.805 & 0.847 & 0.820 & 0.835 \\ 
\multirow{2}{*}[+10pt]{Obj.} & SEPARATE & 51.9M  & {0.859} & {0.876} & {0.902} & {0.915} & {0.818} & {0.854} & {0.838} & {0.869} \\
\bottomrule
\end{tabular}
\caption{Simulation test-set performance for our joint grasp type classifier (COMBINED) and an ensemble of individual networks (SEPARATE). \# Params shows the number of learnable parameters. Average test-set accuracy, precision, and F1 score at convergence are followed by accuracy for all $n=5$ grasp types (wide power, wide precision, basic power, basic precision, pincher).}
\label{tab:simacc}
\end{table*}

Though test-set accuracy demonstrates how well the system learns, when executed on a real robot, selecting false positives can cause a low grasp success rate.
Since the system chooses to execute the one grasp with the highest predicted success probability, the grasp may fail if an incorrectly classified false positive is selected.
False negatives, however, are not as detrimental to the grasp success rate since the system will choose only one of the many grasps expected to be successful.
It is, therefore, also important to verify the system's precision and F1 score.

As COMBINED is trained to predict success for all grasp types, all statistics are reported at the epoch at which average accuracy across all grasp types is maximum; the individual grasp type accuracies are not necessarily maximum at this epoch.
For SEPARATE, since each network is trained with only one grasp type, each accuracy, precision, and F1 score are reported at the epoch at which that grasp type's accuracy is maximum. 
The reported average accuracy is the average of these maximum accuracies.
Despite this, when learning these binary classifiers jointly from a shared embedding, the average classification accuracy decreases by only 0.4\% when test objects have been seen and 3.0\% when the training and test object sets are exclusive.
Furthermore, our system achieves a higher precision in the case where test candidates are selected at random.
These experiments show that jointly training individual classifiers from a shared PointConv embedding enables our system to more efficiently classify grasp poses with a negligible loss in performance compared to a similar set of networks with five times as many parameters.

\section{Clearing a Cluttered Table}
\label{sec:experiments}
\begin{figure}[h]
\vspace{7pt}
\centerline{\includegraphics[scale=0.25]{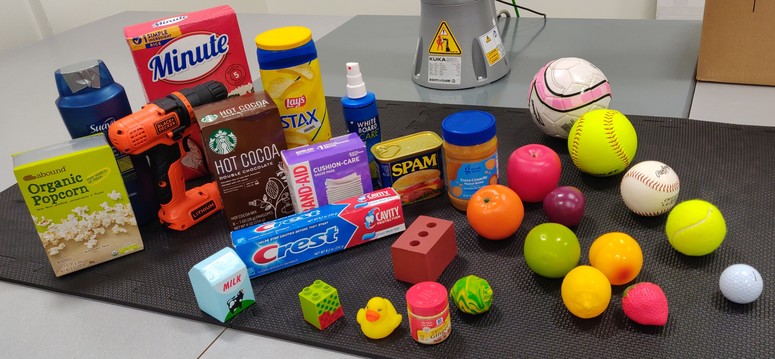}}
\caption{Objects used in real-world experiments, none of which are found in our simulated training set.}
\label{fig:objs}
\end{figure}

We perform real-robot experiments to measure how much multi-modal grasps help to clear objects of varying sizes from a cluttered table.
We capture two point clouds from fixed locations, then remove all points within a threshold of the known table plane.
As described in Section \ref{sec:approach}, we generate a set of 400 grasp candidate poses $G$ using $\mathbf{GEN}$, then check each for collisions, both in PyBullet with a simplified mesh and between the cloud and a simplified gripper model.
Like \citet{rob-journal}, we also filter candidates with an insufficient number of points in the graspable region between the fingers.
This is achieved by counting the number of points in the box each finger would sweep through as the gripper closes.
We then evaluate the remaining candidates with our trained $\mathbf{DNN}$.
As detailed in Section \ref{sec:approach}, we choose to execute grasp $g_m$ of type $i_m$ with predicted success probability $S_m$ using our Robotiq 3-Finger Adaptive Gripper.
If our motion planner fails to find a path to $g_m$, we execute the next most likely to succeed grasp.

Because related works are evaluated with a variety of datasets and often simpler experimental scenarios, and implemented on different robot hardware, it is difficult to compare our system with them directly.
To examine the benefits of multiple grasp types when grasping in dense clutter, we compare our system to two baseline ablations representative of related work whose deep networks have not been trained to assess all five grasp types.
The first, 1Type, predicts only the probability that a pincher grasp would succeed, and is representative of systems designed for parallel-jaw grippers~\citep{rob-journal, dexnet2}.
The second, 2Type, predicts whether $n=2$ grasp types, basic power and basic precision, would succeed; this is representative of the framework defined by \citet{tucker-grasp-types}.
Our system, 5Type, models all $n=5$ Robotiq grasp types.

\begin{figure}[h]
\centerline{\includegraphics[scale=0.25]{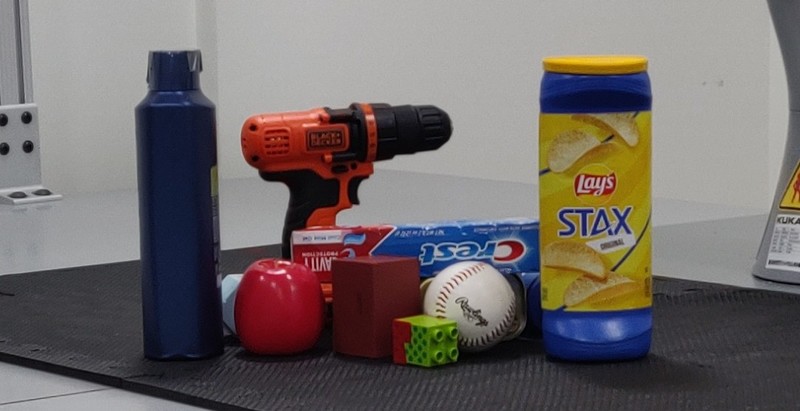}}
\caption{Ten small and medium objects in a cluttered pile surrounded by three large, upright objects.}
\label{fig:exp_setup}
\end{figure}

In each of our experimental trials, three large, upright objects are placed around a pile of ten small and medium-sized objects.
This scenario is designed to challenge the system, as it may depend on all five grasp types to clear the table.
An example of the clutter our system clears is shown in Figure \ref{fig:exp_setup}.
The objects used in these experiments, an augmented segment of the YCB dataset \citep{ycb} containing six large, 17 medium, and six small objects, did not appear in the training set and can be seen in Figure \ref{fig:objs}.

Our procedure follows that of \citet{rob-journal}; a random selection of small and medium objects is placed in a box, shaken, and dumped into a cluttered pile on a table; large items are placed around the pile after dumping the box to ensure they remain upright.
The system attempts to grasp objects until either 1) the same type of failure on the same object with the same grasp type fails three times in a row, 2) the system fails to generate reachable grasps in three consecutive attempts, or 3) all objects are removed from the table.
If the system fails to find a feasible grasp or the motion planner fails to find paths to the top 25 grasps, we repeat the candidate generation process up to two more times, each time proposing twice as many candidates.
A grasp is successful if one or more objects are lifted from the scene and moved towards a box, and do not fall from the gripper until the fingers are opened.
If an object leaves the workspace during an unsuccessful grasp or during a grasp on a different object, it is placed back in the scene near where it left, abutting as many other objects as possible.
If an unforeseen collision occurs during a grasp or placement execution, the disturbed objects are reset and the attempt is not counted.
Each system is presented with the same objects as the other systems in each of the ten trials, but in a different random configuration.
We report both the grasp success rate (number of successful grasps divided by number of attempted grasps) with number of attempts and overall object removal rate (number of objects removed from table at the end of a trial divided by initial number of objects); results are in Table \ref{tab:exp}, and an example trial is shown in the accompanying video.

\begin{table}[h]
\centering
\begin{tabular}{cccc}
\toprule
& Success Rate & \#Attempts & Removal Rate \\
\midrule
\textbf{5Type} & 0.808 & 125 & \textbf{0.808} \\
\textbf{2Type} & 0.692 & 120 & 0.654 \\
\textbf{1Type} & \textbf{0.825} & 114 & 0.723 \\
\bottomrule
\end{tabular}
\caption{Real-world performance on the experimental scenario averaged across ten trials.}
\label{tab:exp}
\end{table}

\section{Discussion and Conclusion}
\label{sec:conclusion}
As seen in Table \ref{tab:exp}, our system with access to all five grasp types, 5Type, outperforms the ablations of our system, 2Type and 1Type, when grasping items from cluttered scenes surrounded by large objects.
5Type achieved the highest object removal rate by successfully clearing nearly all scenes.
In some trials, though $\mathbf{GEN}$ generated grasps on the small Lego or duck, the motion planner failed to find a path to these grasps.
A common failure mode of the system, which is a common failure mode in other grasp detection systems \cite{rob-journal}, was that it attempted to grasp multiple objects at once.
The 5Type system suffers from this issue more than the baselines because it has access to wide-type grasps that spread the fingers out and are more likely to contact multiple cluttered objects.
The issue could be alleviated with an off-the-shelf depth image segmentation algorithm and an additional filtering step in $GEN$.
Other failure modes of our system include false positives and objects moving as the fingers close.
The system also struggled with the heavy shampoo bottle in two of the five trials it appeared in.
Because our grasp evaluation network makes predictions based only on local geometry, it incorrectly predicted that unstable grasps would succeed.
Since our system has no notion of grasp history, it became stuck in a local minima and unsuccessfully attempted similar grasps three times in a row, ending the trials.
However, in three of the trials, it correctly used power grasps to stably lift the heavy bottle.

Since 2Type is incapable of performing pincher grasps, it often fails to find feasible grasp candidates on small objects that fit between the gripper's spread fingers.
The system encountered difficulty with the rice box in one trial and the pear in another, objects that the 5Type system never failed to grasp.
However, because this ablation did not have access to the weak but precise pincher grasps, it was able to successfully grasp the heavy shampoo bottle in three of the five trials it appeared in.
The 1Type system outperformed the 2Type system in these experiments because pincher grasps succeeded on the small or medium objects that made up the majority of our object set.
Even most of the large objects were light enough that they could be lifted with a pincher grasp.
1Type attempted to grasp multiple objects at once less frequently because the fingers are not spread apart during a pincher grasp.
The 1Type system successively failed to lift the heavy shampoo bottle by its cap three times in three different trials, ending the trials prematurely and decreasing the object removal rate.

Our 5Type system was able to choose applicable grasp types for the situation to clear each scene more completely.
It used a wide power grasp to stably grasp the top of the drill, and another to perform a spherical grasp on the soccer ball, one of the largest medium-sized objects.
It used basic power grasps when faced with the heavy shampoo.
The 2Type system also selected basic power grasps for these objects.
Overall, our 5Type system executed three wide power, 33 wide precision, six basic power, 38 basic precision, and 45 pincher grasps.
Collision-free power grasps were rarely generated on the small and medium-sized objects because they lie close to the tabletop.
As the large objects were often partially occluded by the adjacent clutter pile, their larger graspable areas were hidden.
Though our 5Type system chose to use fingertip grasps in the many applicable scenarios, it used power grasps to lift the heavy objects the 1Type system often could not.

The modular nature of our system will allow us to include our proposed grasp success prediction network as a useful module in other systems, such as an object retrieval system~\cite{thao}.
The inputs or outputs of our system could be partially masked to target specific objects or grasp types without dataset regeneration or network retraining, enabling our system to evaluate grasps in a robot's high-level planner.

The simulation experiments presented in Section \ref{sec:sim_exp} show that our system is able to efficiently learn to jointly evaluate multiple grasp types for a given grasp candidate.
Our real-world experiments show that this architecture enables a robot equipped with a multi-finger gripper to more successfully clear a scene of cluttered objects of various sizes from a tabletop than systems that use fewer grasp types.

\section*{Acknowledgments}

This work was supported by the National Science Foundation under grant IIS-1717569 and NSF CAREER Award 1844960 to Konidaris, and by the ONR under the PERISCOPE MURI Contract N00014-17-1-2699.
The authors wish to thank Ben Burchfiel, Ben Abbatematteo, and Barrett Ames for their helpful feedback and support.

\bibliographystyle{IEEEtranN}
\footnotesize
\bibliography{IEEEabrv, grasp}  

\end{document}